\documentclass[final]{arxiv}

\usepackage{times}
\usepackage{epsfig}
\usepackage{graphicx}
\usepackage{amsmath}
\usepackage{amssymb}
\usepackage{tabularx}
\usepackage{subcaption}
\usepackage{booktabs}
\usepackage{datatool}
\usepackage{siunitx}

\usepackage[pagebackref=true,breaklinks=true,colorlinks,bookmarks=false]{hyperref}

\newcommand*\dtlformat[1]{\DTLifnumerical{#1}{\num{#1}}{#1}}
\newcommand*\dtlboldmax[2]{
  \DTLforeach{#1}{}{
    \def\theMax{0}
    \DTLforeachkeyinrow{\thisValue}{
      \ifthenelse{\dtlcol>#2}{
        \DTLmax{\theMax}{\theMax}{\thisValue}
      }{}
    }
    \DTLforeachkeyinrow{\thisValue}{
      \ifthenelse{\dtlcol>#2}{
        \ifthenelse{\DTLisieq{\thisValue}{\theMax}}{
          \DTLreplaceentryforrow{\dtlkey}{\textbf{\theMax}}
        }{}
      }{}
    }
  }
}

\DTLloaddb{meta_dataset}{data/meta_dataset.csv}
\dtlboldmax{meta_dataset}{2}
\DTLloaddb{meta_dataset_test}{data/meta_dataset_test.csv}
\DTLloaddb{imagenet}{data/imagenet.csv}
\DTLloaddb{imagenet_iid_combinations}{data/imagenet_iid_combinations.csv}
\DTLloaddb{clean_corrupt_error}{data/clean_corrupt_error.csv}
\DTLloaddb{data_aug_corruptions}{data/data_aug_corruptions.csv}

\def\authorSpace{0.805em}

\begin{document}
\begin{filecontents*}{data/meta_dataset.csv}
DATASOURCE,PREPROCESSING,SUR,BB3,BB5,BB7,BA3,BA5,BA7,ERF3,ERF5,ERF7,STRIDE1
Imagenet,$\downarrow\downarrow\downarrow$,43.31,46.04,46.26,45.56,42.49,41.43,40.76,41.92,40.88,40.75,46.40
Omniglot,$\downarrow$,97.28,97.29,97.62,97.17,97.00,96.71,97.21,96.99,96.90,96.94,97.11
Aircraft,$\downarrow\downarrow\downarrow$,90.11,89.60,90.07,89.35,89.74,89.26,87.62,87.18,87.34,86.58,90.88
Birds,$\downarrow\downarrow\downarrow$,70.87,73.98,71.79,70.29,68.93,68.10,68.34,66.10,65.25,65.50,75.41
Textures,$\downarrow\downarrow\downarrow$,66.98,69.05,68.79,66.58,64.54,65.33,65.09,64.13,61.67,58.77,71.17
Quick Draw,$\uparrow\uparrow$,81.66,82.64,82.46,83.24,81.90,81.66,81.82,82.11,81.23,81.03,82.39
Fungi,$\downarrow\downarrow\downarrow$,65.84,68.56,66.57,66.11,64.25,63.08,62.38,62.25,61.83,62.32,70.95
VGG Flower,$\downarrow\downarrow\downarrow$,86.08,88.55,88.46,86.56,84.58,84.71,83.52,84.61,84.21,83.05,88.60
Average accuracy, ,75.27,76.96,76.50,75.61,74.18,73.79,73.34,73.16,72.41,71.87,77.86
\end{filecontents*}

\begin{filecontents*}{data/meta_dataset_test.csv}
DATASOURCE,SUR,SUR-S,AAS3,AAS3-S,AAS5,AAS5-S,AAS7,AAS7-S,GELU,GELU-S,AASUBG3,AASUBG3-S,AASUBG5,AASUBG5-S,AASUBG7,AASUBG7-S,AASG3,AASG3-S,AASG5,AASG5-S,AASG7,AASG7-S
Imagenet,53.77,1.10,57.32,1.13,58.46,1.08,58.67,1.09,56.81,1.11,59.59,1.08,59.37,1.02,59.06,1.02,59.91,1.06,\textbf{60.59},\textbf{1.04},58.03,1.07
Omniglot,95.81,0.36,96.11,0.36,95.87,0.36,\textbf{96.45},\textbf{0.33},96.29,0.32,96.35,0.32,96.39,0.32,96.25,0.34,96.06,0.33,\textbf{96.45},\textbf{0.31},96.38,0.34
Aircraft,87.47,0.49,89.22,0.43,89.54,0.45,89.90,0.46,88.36,0.56,\textbf{90.93},\textbf{0.40},90.21,0.47,89.98,0.48,90.76,0.46,90.34,0.50,90.68,0.45
Birds,72.44,0.98,78.70,0.87,77.86,0.82,77.36,0.90,72.81,0.89,79.52,0.81,\textbf{80.50},\textbf{0.77},79.76,0.73,78.52,0.81,79.96,0.79,79.51,0.80
Textures,68.96,0.77,71.29,0.90,72.08,0.78,72.77,0.92,72.59,0.80,\textbf{75.00},\textbf{0.75},74.36,0.70,74.55,0.78,74.40,0.73,74.98,0.80,72.61,0.77
QuickDraw,81.58,0.60,82.30,0.54,82.29,0.56,82.29,0.59,83.05,0.54,83.81,0.53,83.23,0.56,83.58,0.52,\textbf{84.01},\textbf{0.54},83.18,0.53,82.88,0.57
Fungi,65.67,1.01,69.87,0.93,71.07,0.98,70.48,0.93,67.91,1.03,72.99,0.95,\textbf{73.86},\textbf{0.92},73.57,0.90,72.88,0.92,73.50,0.89,72.98,0.91
VGG Flower,87.61,0.63,89.04,0.55,87.69,0.60,89.25,0.58,87.55,0.58,90.55,0.47,\textbf{91.24},\textbf{0.47},90.31,0.50,88.81,0.52,89.31,0.48,90.09,0.50
Traffic Signs,51.75,1.06,51.05,1.09,\textbf{55.52},\textbf{1.03},52.15,1.07,53.51,1.06,54.17,1.00,51.00,1.02,51.96,0.99,51.55,1.01,53.51,1.04,49.11,1.00
MSCOCO,48.95,1.10,48.30,1.09,49.61,1.05,47.34,1.09,49.71,1.05,49.53,1.05,49.55,1.01,49.38,1.02,50.28,1.04,\textbf{50.72},\textbf{1.02},47.31,1.02
MNIST,94.04,0.49,92.90,0.55,92.11,0.56,88.06,0.51,\textbf{95.42},\textbf{0.43},94.13,0.46,89.77,0.50,94.05,0.47,93.84,0.51,91.02,0.45,92.46,0.44
CIFAR10,62.45,0.91,66.22,0.84,66.06,0.85,62.32,0.93,63.20,1.01,68.45,0.74,68.21,0.78,66.44,0.77,\textbf{69.60},\textbf{0.77},69.29,0.71,63.75,0.80
CIFAR100,53.12,1.10,56.80,1.04,57.03,1.05,53.63,1.09,56.81,1.21,57.67,1.03,58.50,1.06,58.22,1.02,\textbf{61.58},\textbf{1.04},59.61,1.02,53.40,1.04
Average,71.05,-,73.02,-,73.48,-,72.36,-,72.62,-,\textbf{74.82},-,74.32,-,74.39,-,74.79,-,\textbf{74.80},-,73.01,-
Average {\tiny(in-domain)},76.66,-,79.23,-,79.37,-,79.65,-,78.17,-,81.09,-,\textbf{81.15},-,80.88,-,80.67,-,81.04,-,80.40,-
Average {\tiny(out-of-domain)},62.06,-,63.07,-,64.07,-,60.70,-,63.73,-,\textbf{64.79},-,63.41,-,64.01,-,65.37,-,\textbf{64.83},-,61.21,-
\end{filecontents*}

\begin{filecontents*}{data/imagenet.csv}
WHERE,BB3,BB5,BB7,BA3,BA5,BA7,BBOTH3,BBOTH5,BBOTH7,ERF3,ERF5,ERF7
conv all layers,       74.20,71.06,68.74,73.91,70.71,68.00,70.92,65.80,61.52,73.06, 71.09,68.67
conv1,     76.18,75.81,75.45,76.10,75.72,75.47,75.76,75.07,74.63,75.27, 75.12,75.03
max pool,  76.00,75.75,75.40,76.56,76.43,76.35,76.03,75.54,75.24,74.88, 74.87,74.60
block-conv,76.87,76.74,76.83,76.88,76.83,76.74,76.87,76.73,76.63,76.64, 76.71,76.61
skip,      77.05,77.14,77.07,\bf{77.15},77.12,76.83,77.02,76.90,76.62,-,-,-
\end{filecontents*}

\begin{filecontents*}{data/imagenet_iid_combinations.csv}
Method,BB3,BB5,BB7,BA3,BA5,BA7
Subsampled, , , , , , 
Subsampled (low capacity), , , , , , 
\end{filecontents*}

\begin{filecontents*}{data/clean_corrupt_error.csv}
Method,SKIP,MAX-POOL,BLOCK-CONV,CLEAN,GAUSS,SHOT,IMPULSE,DEFOCUS,GLASS,MOTION,ZOOM,SNOW,FROST,FOG,BRIGHT,CONTRAST,ELASTIC,PIXEL,JPEG,MCE
ResNet-50 \tiny{Published} \cite{hendrycks2019robustness}, , , ,23.9,80,82,83,75,89,78,80,78,75,66,57,71,85,77,77,76.7
ResNet-50 \tiny{Ours}, , , , 23.5,69,71,71,73,88,80,78,73,69,50,54,66,82,80,67,71.4
Ours,\checkmark, , ,22.9,68,70,70,72,87,78,75,73,69,50,53,66,81,84,68,70.9
Ours, ,\checkmark, ,23.4,70,72,73,72,87,80,79,72,69,51,54,67,83,84,68,72.0
Ours, , ,\checkmark,23.2,70,72,72,72,88,80,77,73,69,50,53,66,81,84,67,71.6
Ours,\checkmark, ,\checkmark, \textbf{22.5},69,71,72,72,87,79,76,72,68,50,52,64,81,84,68,70.9
Ours,\checkmark,\checkmark, ,22.9,70,71,73,72,80,79,75,71,67,51,53,67,82,85,68,71.2
% anti-alias sub-sampling combined
Ours,\checkmark,\checkmark,\checkmark,\textbf{22.5},68,70,70,72,85,82,75,72,62,50,52,66,81,81,66,\textbf{70.0}
BlurPool \cite{zhang2019shiftinvar},\checkmark,\checkmark,\checkmark,23.0,73,74,76,74,86,78,77,77,72,63,56,68,86,71,71,73.4
\end{filecontents*}

\begin{filecontents*}{data/data_aug_corruptions.csv}
Method,SKIP,MAX-POOL,BLOCK-CONV,CLEAN,GAUSS,SHOT,IMPULSE,DEFOCUS,GLASS,MOTION,ZOOM,SNOW,FROST,FOG,BRIGHT,CONTRAST,ELASTIC,PIXEL,JPEG,MCE
AutoAugment**\cite{DBLP:journals/corr/abs-1805-09501}, , , ,22.8,69,68,72,77,83,80,81,79,75,64,56,70,88,57,71,72.7
Rand AutoAugm.**\cite{hendrycks2020augmix}, , , ,23.6,70,71,72,80,86,82,81,81,77,72,61,75,88,73,72,76.1
BlurPool \cite{zhang2019shiftinvar}, , , ,23.0,73,74,76,74,86,78,77,77,72,63,56,68,86,71,71,73.4
AUGMIX \cite{hendrycks2020augmix}, , , ,22.4,65,66,67,70,80,66,66,75,72,67,58,58,79,69,69,68.4
ANT (3x3) \cite{rusak2020simple}, , , ,23.9,39, 40, 39, 68, 78, 73,77,71,66,68,55, 69,79,63,64, 63.0
Baseline, , , ,22.6,70,72,72,71,87,79,76,73,69,51,53,66,81,81,67,71.2
RandAugment** \cite{cubuk2019randaugment}, , , ,22.8,60,58,60,70,90,76,80,70,67,44,50,57,80,86,64,67.4
Ours {\tiny + RandAugm.**},\checkmark, , ,21.8, 60, 60, 60, 70 , 85, 72, 76,69 , 65, 42, 48 , 55 , 80, 86, 62 ,66.2
Ours {\tiny + RandAugm.**},\checkmark,\checkmark,,21.5, 58, 57, 59, 70, 85, 72, 76 , 69, 66, 42, 48 ,56, 78, 82, 62 ,65.2
Ours {\tiny + RandAugm.**},\checkmark, ,\checkmark,21.5, 58, 58, 59, 70, 86, 74, 75,69 , 64, 41,47 ,55 , 79, 83,62,65.4
Ours {\tiny + RandAugm.**},\checkmark,\checkmark,\checkmark,21.6, 59,  58, 61, 70, 84 , 75, 76, 69, 65, 41, 48,55, 80, 82,61, 65.5
Swish, , , ,22.4,71,72,74,69,88,80,76,74,69,51,54,68,81,80,67,71.6
Ours {\tiny+ Swish + Rand Augm.}, , , ,21.9,61,61,62,69,88,73,78,69,67,42,49,55,81,87,63,66.9
Ours {\tiny+ Swish + Rand Augm.},\checkmark, , ,21.4,59,59,59,69,85,74,76,59,65,42,47,55,78,79,62,65.1
Ours {\tiny+ Swish + Rand Augm.},\checkmark,\checkmark, ,21.4,60,59,60,69,85,73,76,67,65,42,47,55,78,82,62,65.3
Ours {\tiny+ Swish + Rand Augm.},\checkmark, ,\checkmark,21.1,60,60,63,69,86,71,75,68,64,41,47,55,78,83,62,65.3
Ours {\tiny+ Swish + Rand Augm.},\checkmark,\checkmark,\checkmark,21.2,60,59,61,69,85,71,75,68,64,41,47,55,78,81,61,64.9
\end{filecontents*}

\title{An Effective Anti-Aliasing Approach for Residual Networks}

\author{

{Cristina Vasconcelos \hspace{\authorSpace} Hugo Larochelle \hspace{\authorSpace} Vincent Dumoulin \hspace{\authorSpace} Nicolas Le Roux \hspace{\authorSpace} Ross Goroshin}\\
\and 
Google Research, Montr\'{e}al \\
{\tt\small \{crisnv, hugolarochelle, vdumoulin, nlr, goroshin\}@google.com}
}

\maketitle


\begin{abstract}
Image pre-processing in the frequency domain has traditionally played a vital role in computer vision and was even part of the standard pipeline in the early days of deep learning. However, with the advent of large datasets, many practitioners concluded that this was unnecessary due to the belief that these priors can be learned from the data itself. Frequency aliasing is a phenomenon that may occur when sub-sampling any signal, such as an image or feature map, causing distortion in the sub-sampled output. We show that we can mitigate this effect by placing non-trainable blur filters and using smooth activation functions at key locations, particularly where networks lack the capacity to learn them. These simple architectural changes lead to substantial improvements in out-of-distribution generalization on both image classification under natural corruptions on ImageNet-C~\cite{hendrycks2019robustness} and few-shot learning on Meta-Dataset~\cite{triantafillou2020metadataset}, without introducing additional trainable parameters and using the default hyper-parameters of open source codebases. 
\end{abstract}

\section{Introduction}
Deep learning approaches thrive in problem settings where labelled data is abundant and training times are virtually unrestricted – allowing the algorithm to apparently learn all necessary features and priors to achieve robust performance. Central to the success of deep learning approaches on supervised learning problems is the assumption that the training and test data are sampled from the same distribution. However, many important problem settings involve out-of-distribution (OOD) data or may restrict the amount of labelled data typically required for training deep networks ``from scratch''. In this work we investigate two such settings, namely: image classification under natural corruptions and few-shot classification. In these challenging scenarios, where implicit knowledge cannot be fully obtained from the training datasets, the search for stronger architectural priors as a mechanism to impose explicit knowledge is a promising line of investigation.

\begin{figure}
\centering
\hfill
\includegraphics[width=0.5\linewidth]{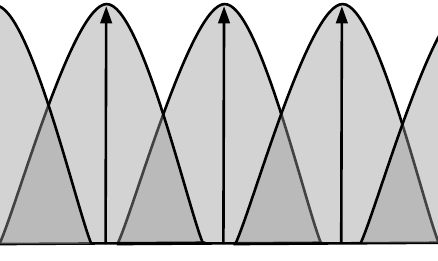}
\hfill
\includegraphics[width=0.45\linewidth]{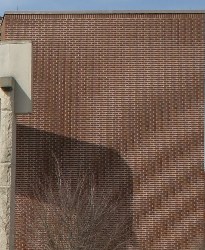}
\hfill
\caption{{\em Left}: Spectral aliasing illustrated in the frequency domain, showing the effect on the spectrum of a sub-sampled signal. The overlapping dark gray regions represent aliased frequencies {\em Right}: An  image with aliasing distortion \cite{wiki1}.}
\label{fig:aliasing}
\end{figure}

\section{Spectral Aliasing in Convolutional Networks}
Spectral aliasing is a phenomenon that may occur when sub-sampling any signal, such as an image or feature map. 
Aliasing occurs when the sampling rate is too low for the bandwidth of the signal and does not satisfy the Nyquist rate~\cite{openheim1997signals}. When a signal is aliased, its high frequency components additively spill into the low frequency components. This type of corruption can result in noticeable artifacts, such as the ripple effect (Moir\'e pattern) observed over the brick wall depicted in \autoref{fig:aliasing}. In classical image/signal processing, aliasing is prevented by applying an {\em anti-aliasing} filter before sub-sampling. The ideal low-pass filter, the ``sinc'' function, is usually replaced by non-ideal low pass filters such as the Gaussian, for computational efficiency.

In convolutional networks, aliasing may be caused by any operation that spatially sub-samples the input, including pooling and convolutional layers with stride greater than one. As a result, subsequent layers of the network may learn features that rely on the presence (or absence) of aliasing artifacts. This may affect generalization in settings where test images have different spectral properties, such as images of varying size, compression format, or images affected by natural corruptions.

In this work, we attempt to answer the following questions:
\begin{itemize}
    \item {\it Does frequency aliasing affect the performance of convolutional networks on challenging tasks such as few-shot learning and classification under natural corruptions?} Although input or feature map aliasing may occur in deep networks, it is not clear if this phenomenon affects their performance on highly abstract, semantic tasks.
    \item {\it Do convolutional networks implicitly learn to prevent aliasing?} It can be shown that a sequence of convolutions can be represented as a single convolution with a larger kernel. Therefore, it is plausible that anti-aliasing filters can be learned by existing, trainable filters of convolutional networks provided that the filters provide sufficient spatial support. 
    \item {\it Does the placement of anti-aliasing filters influence performance?} Theoretically, anti-aliasing filters can be placed anywhere in the network.
    However our experiments show that the placement of anti-aliasing filters is critical for achieving good performance. We show that the ResNet~\cite{DBLP:conf/cvpr/HeZRS16,He2016} architecture lacks the capacity to learn anti-aliasing filters in some of its pathways. More specifically, we experimentally verify that aliasing effects are most severe in the skip-connection paths that include sub-sampling. These paths typically include only $1 \times 1$ convolutions, which lack the capacity to learn anti-aliasing filters by construction. Such pathways will produce aliasing artifacts whenever they receive non band-limited features as input. 
    \item {\it Are networks incentivized to learn anti-aliasing filters?} Even if networks have the capacity to learn anti-aliasing filters, they may not be trained with the required data augmentation, a rich enough dataset, or specific regularizer, that would incentivize them to learn features that are invariant to aliasing artifacts. Furthermore, because anti-aliasing filters are a simple and robust solution, it would be extremely sample inefficient to learn these. 
    
    \item {\it Can we separate aliasing from other confounding effects?} Introducing anti-aliasing filters can affect both inference and training. We perform a number of ablation studies that aim to isolate the effects of aliasing from other possible influences on performance, including potential increase in receptive field size, and interaction of the anti-aliasing filter with backpropagation dynamics, as it may alter the spatial neighborhood that influences the computation of gradients.
\end{itemize}
Our results show that the architectural modifications, detailed in \autoref{fig:best_model}, i.e.\ blur applied after(1) strided-skip connections, (2) strided-convolutional layers from the residual blocks main path, and (3) initial strided-max pooling, combined with smooth activations, and data augmentation achieves impressive performance on several datasets. It surpasses the state-of-the-art stand-alone method for Imagenet-C on 9 of 15 categories, while also producing the lowest clean error on Imagenet, and improves further the performance of one of the published state-of-the-art methods on Meta-Dataset. 
We emphasize that it is particularly important to add anti-aliasing filters to the pathways that lack the capacity to learn them. 

\section{Related Work}
Zhang~\cite{zhang2019shiftinvar} observe that max-pooling operations can be decomposed into two operations: evaluating the max operation densely, followed by sub-sampling; and argue that a similar decomposition can be applied to strided convolutional layers and average pooling layers. In contrast to our approach, they propose the use of a blur filter after the sub-sampling's layer non-linearity. In their model, a strided-convolution followed by ReLU activation is redefined as non-strided convolution followed by ReLU activation, followed by a strided-blur. The blurring operation also smooths the output of the non-linearity. This not only prevents the output of the non-linearity from producing high-frequency activations, but also does not fully isolate the aliasing phenomenon. On the other hand, our model postpones removing any high frequencies caused by the non-linearity up to the subsequent sub-sampling operation. 

Azulay and Weiss~\cite{DBLP:journals/jmlr/AzulayW19} point out that convolutional networks are not as robust to small image transformations as commonly assumed. They also mention that convolutional networks typically ignore the sampling theorem and report large changes in the output of modern architectures under small, mostly imperceptible, perturbations of the input. They also argue that the improvement in the generalization obtained by data augmentation is limited to images that are similar to those seen during training. 
%

Zou et al.~\cite{zou2020delving} propose the use of trainable low-pass filtering layers that operate on feature channel groups and adapt to spatial locations. Their anti-aliasing module consists of trainable components and extra non-linearities, increasing the capacity of the network, thus making it unsuitable for isolating the effects of aliasing as is done in our work.

\begin{figure*}
\begin{center}
\includegraphics[width=0.9\textwidth]{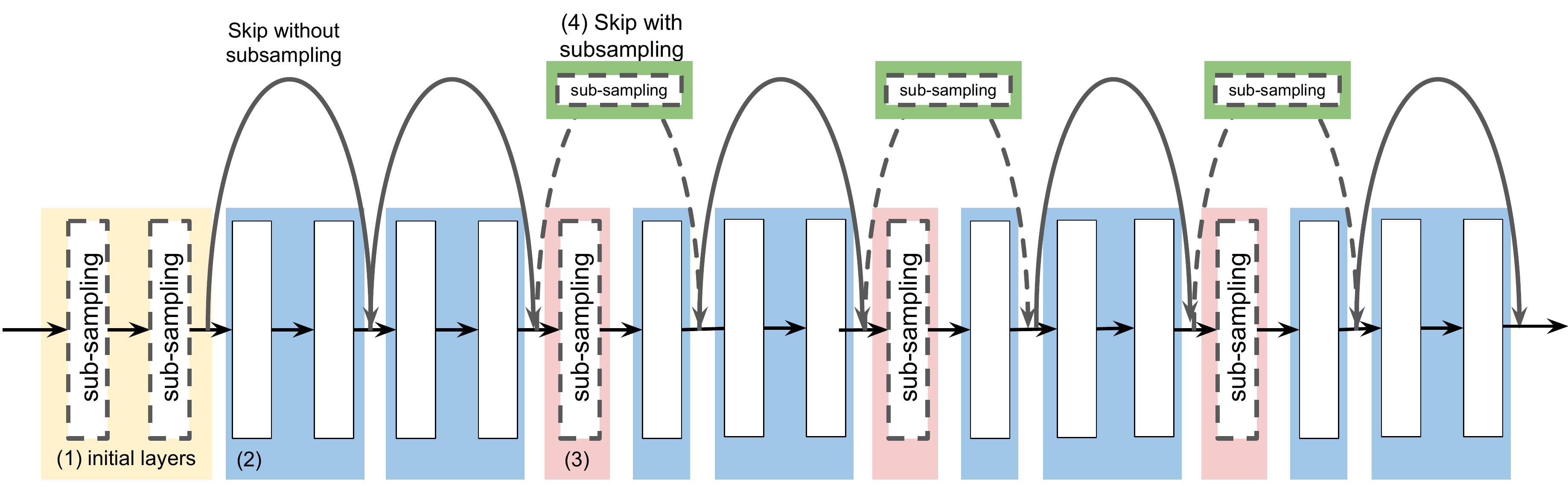}
\end{center}
   \caption{We investigate incorporating anti-aliasing operations at various locations in the ResNet architecture: (1) in the initial layers; in the main path of residual blocks (2) without sub-sampling and (3) with sub-sampling; and (4) in the skip connections with sub-sampling. Dashed lines indicate modules with sub-sampling (stride 2).}
\label{fig:modules_labels}
\end{figure*}

\begin{figure}[ht]
\centering
\includegraphics[width=0.31\linewidth]{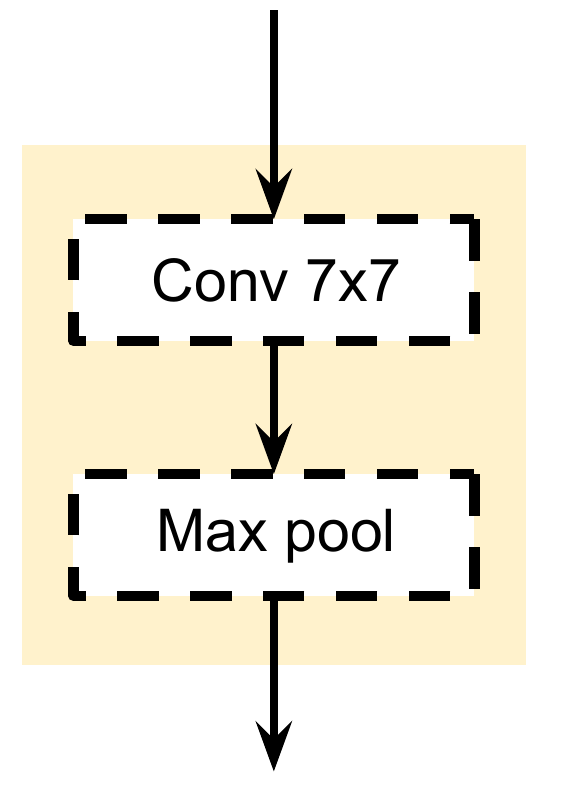}
\hfill
\includegraphics[width=0.68\linewidth]{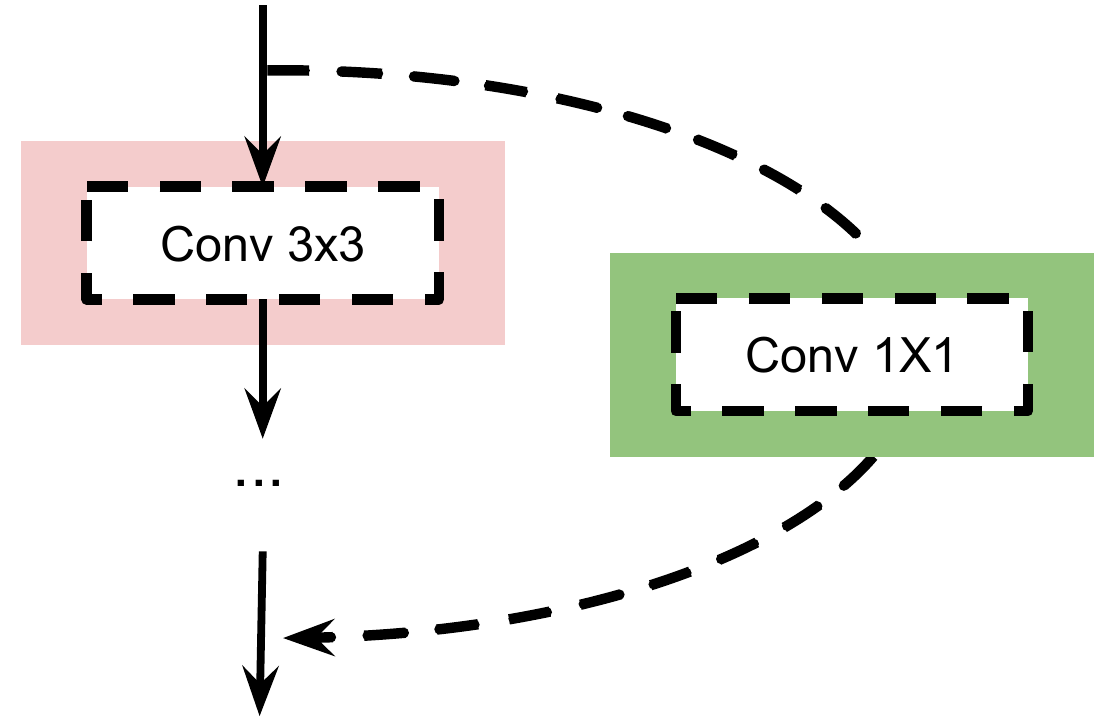}
\caption{A closer look at the ResNets components that include sub-sampling. Dashed lines indicate modules with sub-sampling (stride 2). Left: Initial layer (yellow) consist of convolutions with large trainable kernels, capable of learning anti-aliasing filters. Right: skip connection filters (1 $\times$ 1) lack the spatial support to serve as anti-aliasing filters on their input.
Non-linearities and layers that maintain spatial resolution are omitted for clarity.}
\label{fig:closer_look}
\end{figure}

\begin{figure}[ht]
\centering
\includegraphics[width=\linewidth]{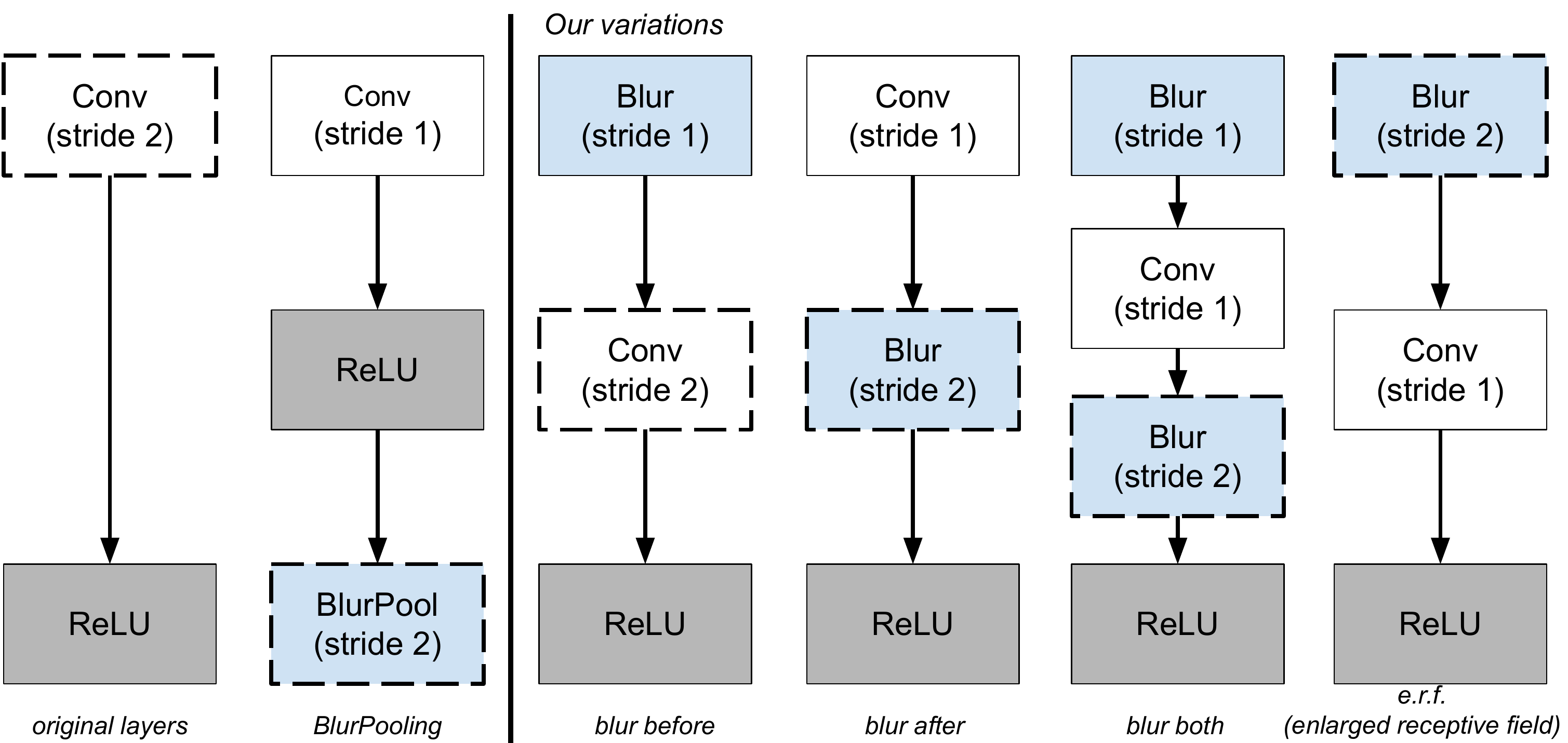}
\caption{\label{fig:blur_variations}
Anti-aliasing can be added at various locations with potentially different side-effects. {\em BlurPooling}~\cite{zhang2019shiftinvar} performs anti-aliasing followed by sub-sampling after the ReLU non-linearity, whereas our variants 
perform anti-aliasing around the convolution operation and before the non-linearity. The enlarged receptive field ({\em ERF}) variant only applies when the kernel of the original convolutional layer has spatial support larger than 1.}
\end{figure}

Sophisticated augmentation strategies are currently the state-of-the-art approach to OOD classification under natural corruptions~\cite{DBLP:journals/corr/abs-1805-09501,hendrycks2019robustness,hendrycks2020augmix,rusak2020simple}.
Rusak et al.~\cite{rusak2020simple} currently lead the ``standalone leaderboard'' on ImageNet-C with a two pass approach. First, they train a generative model to produce additive noise. Next, the classifier and the generative network are adversarially-trained jointly. Recently, \cite{NEURIPS2019_b05b57f6} pointed out that the obtained improvement is mainly on corruptions that affect high frequencies, while reducing robustness to corruptions that affect low frequencies. Their proposed final classifier has its robustness reduced to both low frequency corruptions and ``clean'' (uncorrupted) test error.

While \cite{NEURIPS2019_b05b57f6} pointed out that robustness gains are typically non uniform across corruption types and that increasing performance in the presence of random noise is often met with reduced performance in corruptions concentrated in different bandwidths, our method overcomes this trade-off. Note that severe aliasing may affect the entire spectrum, and not only high frequencies. 

Our results, combining our architectural modifications with off-the-shelf data augmentation (Ekin et al.~\cite{cubuk2019randaugment}) show improvements in all of the 15 corruption categories and outperforms \cite{rusak2020simple} in 9 of them (including fog and contrast, that are concentrated in low frequencies according to \cite{NEURIPS2019_b05b57f6}) while obtaining the highest clean accuracy of the benchmark: $78.8\%$ for Imagenet versus $76.1\%$ for \cite{rusak2020simple}. At time of writing, we are not aware of better performing standalone methods on Imagenet-C using a Resnet-50 trained on Imagenet only (on 224$\times$224 resolution).

\emph{In contrast to the above methods that settle for a trade-off between clean and corrupted accuracy, our method improves both IID and OOD accuracy simultaneously.} 

Our contributions are summarized as follows: (i) a detailed study aimed at more precisely isolating the effects of aliasing in convolutional networks (see \autoref{fig:blur_variations}) is presented. (ii) A new, simple, architectural modification to ResNet models that does not increase the number of trainable parameters is proposed. (iii) We show that this architectural improvement is complementary to others, namely smooth activations and data-augmentation. (iv) We show that our proposed architecture leads to improved performance on two challenging OOD benchmarks, using open source codebases, and their default hyper-parameters.

\section{Description of Experiments}

\begin{figure*}[t]
\centering
\captionsetup[subfigure]{labelformat=empty}
\begin{minipage}{0.4\textwidth}
\centering
\includegraphics[width=\textwidth]{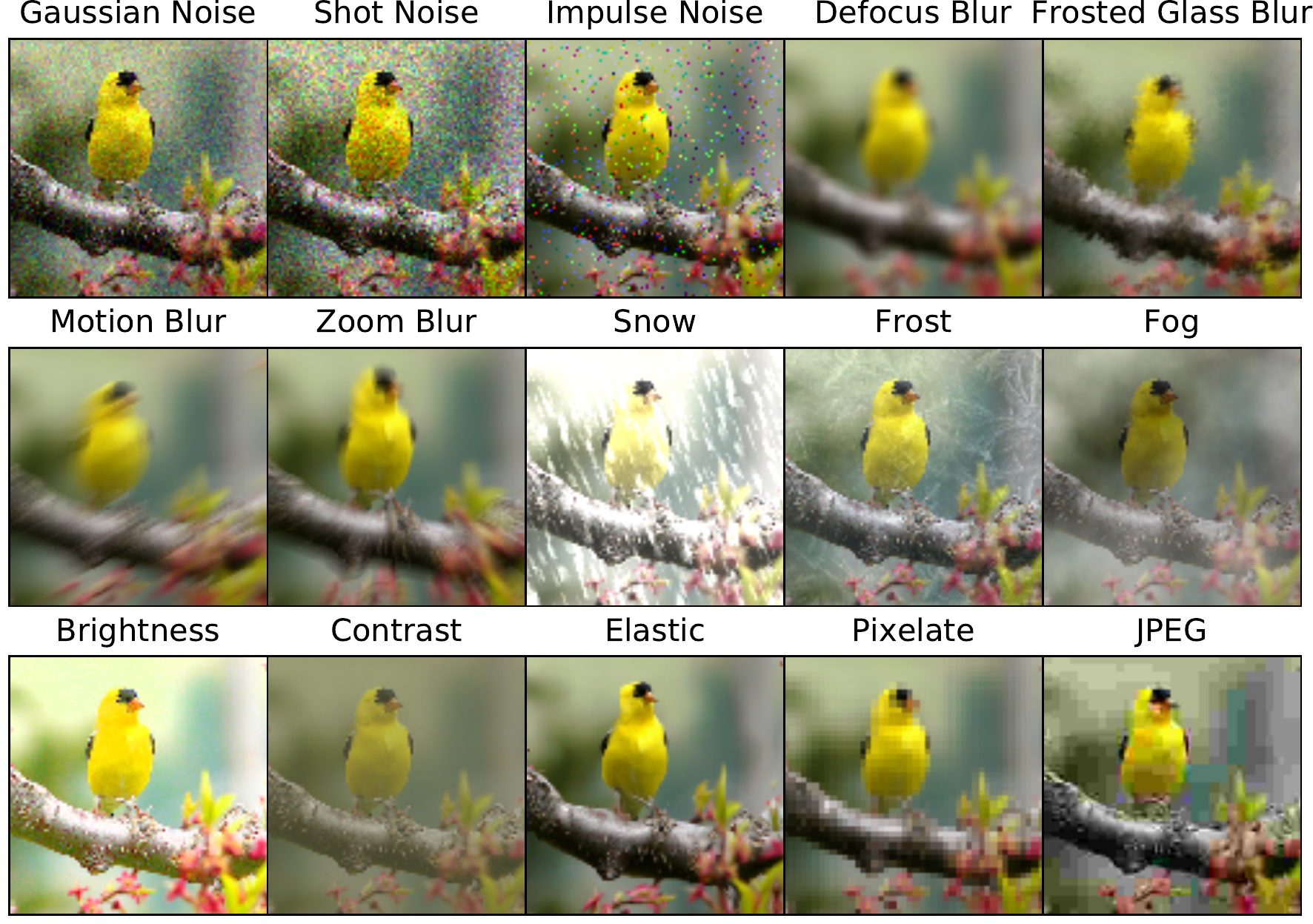}
\end{minipage}
\hfill
\begin{minipage}{0.59\textwidth}
\centering
\subfloat[\footnotesize ImageNet]{\includegraphics[width=0.19\textwidth]{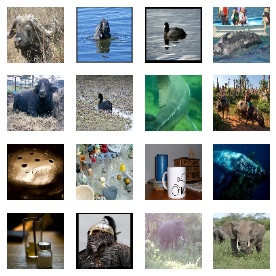}}\hfill
\subfloat[\footnotesize Omniglot]{\includegraphics[width=0.19\textwidth]{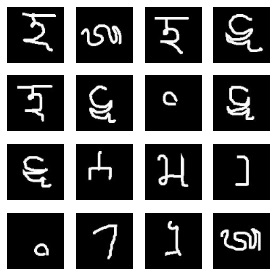}}\hfill
\subfloat[\footnotesize Aircraft]{\includegraphics[width=0.19\textwidth]{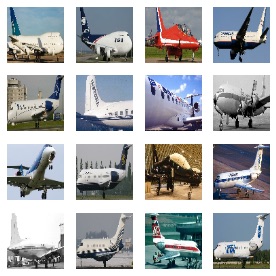}}\hfill
\subfloat[\footnotesize Birds]{\includegraphics[width=0.19\textwidth]{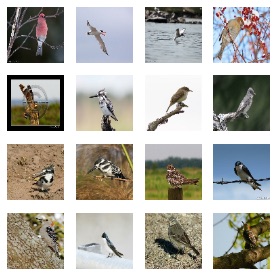}}\hfill
\subfloat[\footnotesize Textures]{\includegraphics[width=0.19\textwidth]{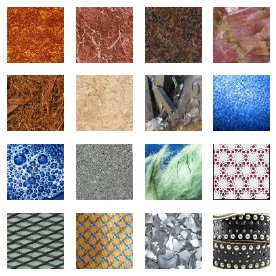}}\\
\subfloat[\footnotesize QuickDraw]{\includegraphics[width=0.19\textwidth]{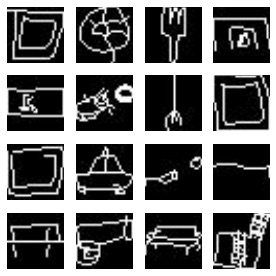}}\hfill
\subfloat[\footnotesize Fungi]{\includegraphics[width=0.19\textwidth]{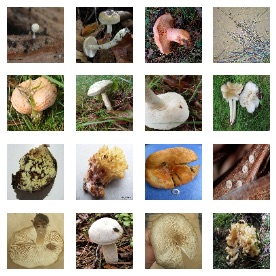}}\hfill
\subfloat[\footnotesize Flower]{\includegraphics[width=0.19\textwidth]{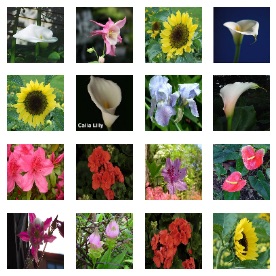}}\hfill
\subfloat[\footnotesize Traffic Signs]{\includegraphics[width=0.19\textwidth]{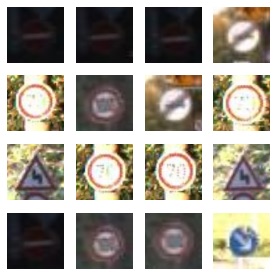}}\hfill
\subfloat[\footnotesize MSCOCO]{\includegraphics[width=0.19\textwidth]{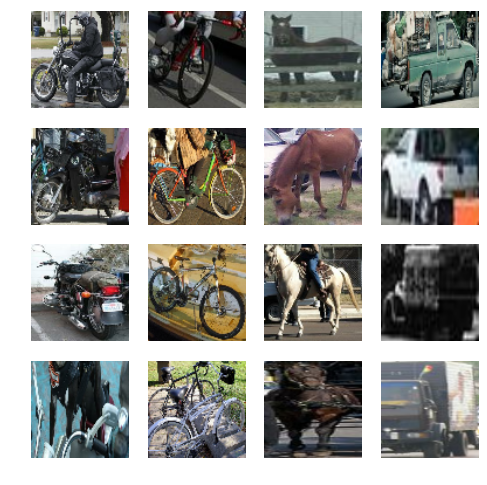}}
\end{minipage}
\caption{\label{fig:datasets}{\em Left}:  ImageNet-C corruptions. {\em Right}: Samples from all 10 data sources included in the Meta-Dataset benchmark.
Figures taken from  Hendrycks and Dietterich~\cite{hendrycks2019robustness} and Triantafillou et al.~\cite{triantafillou2020metadataset}, respectively.}
\end{figure*}

We investigate the effects of sub-sampling and anti-aliasing operations when using the ResNet ~\cite{DBLP:conf/cvpr/HeZRS16,He2016} architecture. We partition various components of the ResNet model into the following ``modules'' (note that each module may include convolutions, non-linearities, and batch normalization): (1) a module representing the initial layers; (2) a  module representing  residual blocks main-path without spatial sub-sampling and (3) one for those with sub-sampling; and (4) a module representing skip-connections with spatial sub-sampling. \autoref{fig:modules_labels} illustrates the partition on a ResNet-18 model. This modular representation will be used to describe various ResNet architectures in our experiments. \autoref{fig:closer_look} takes a closer look at modules with sub-sampling and points out that their filter sizes differentiate them in their capacity to prevent aliasing of the feature maps that immediately precede sub-sampling.

Furthermore, we study several anti-aliasing filter placement variants, shown in \autoref{fig:blur_variations}, to isolate potential side-effects on two aspects of training, namely: (1) on back-propagation dynamics of the trainable convolutional filter 
and (2) on the interaction with the sub-sampling operation. 
All of our anti-aliasing variants prescribe inserting a non-trainable filter in sequence with an ordinary trainable convolutional layer. Two convolutions \emph{per se} are commutative (see \cite{book_097456074X} for a detailed proof). 
However, we observe that this property no longer holds when back-propagating gradients or when the layers include sub-sampling. That is, a change in the order of the fixed and trainable filters induce different priors. 
For example, the \emph{backwards} pass in our \emph{blur after} variant induces the trainable convolution filter to be influenced by a larger neighborhood of gradients than in our \emph{blur before} variant.
Similarly, the decision of where to include sub-sampling affects the \emph{resolution} of the feature maps and also of the gradients that influence the trainable layer.   In our variant named \emph{enlarged receptive field (e.r.f.)}, we explore the special case where sub-sampling is applied at a blur filter that precedes a trainable filter, with a potential side effect of altering the trainable filter receptive field for filters with spatial size ($k$) larger than one.

Our experiments examine both the effects of inserting anti-aliasing filters into the various combinations of the ResNet modules and the different anti-aliasing variants proposed.


We run experiments on top of TensorFlow's official ResNet-50 model for ImageNet IID experiments \footnote{\scriptsize{\url{https://github.com/tensorflow/models/tree/master/official/vision/image\_classification/resnet}}}.
The experiments using data augmentation, were built on top of the public source code from \cite{cubuk2019randaugment}. For the OOD experiments in the Meta-Dataset benchmark, we run experiments on top of the publicly available SUR codebase ~\cite{dvornik2020selecting} -- a recent state-of-the-art few-shot classification model.




Finally, we evaluate the impact of anti-aliasing relative to other architectural changes proposed for OOD problems, such as smooth activation functions \cite{Xie_2020_SAT}. We show that their impact is complementary and combining them produces the best results. 
\subsection{ImageNet-C: Robustness to Natural Corruptions}
ImageNet-C \cite{hendrycks2019robustness} is a dataset used for evaluating the robustness of classifiers under natural image corruptions. It consists of the ImageNet validation set corrupted with 15 (plus four optional) types of natural corruptions under various severity levels (\autoref{fig:datasets} depicts ImageNet-C examples). These corruptions distort the distribution of the image spectra to varying degrees. In contrast to previously proposed methods \cite{hendrycks2019robustness, hendrycks2020augmix, rusak2020simple}, which achieve increased robustness (OOD) at the cost of reducing IID performance (i.e. ImageNet validation without corruptions), we demonstrate that our method is the first to achieve state-of-the-art robustness without compromising accuracy on IID performance. We also explore adding sub-sampling layers throughout the architecture, even to those operations disassociated with sub-sampling, and show that this strategy negatively impacts robustness. An advanced data augmentation method, AugMix \cite{hendrycks2020augmix}, recently achieved state of the art results on ImageNet-C. We also evaluate the impact of training with additional data augmentation, namely AutoAugment \cite{DBLP:journals/corr/abs-1805-09501} and RandAugment \cite{cubuk2019randaugment}. We show that our proposed architecture is complementary to data-augmentation and helps to achieve new state-of-the-art results on ImageNet-C. This result also suggests that anti-aliasing cannot be fully learned using existing augmentation strategies.  




\subsection{Few-shot classification, Meta-Dataset, and SUR}

The objective behind few-shot classification is to create models which can learn on new problems with only a handful of labeled training examples. The evaluation procedure it prescribes is to form test {\em episodes} by sub-sampling classes from a held-out set of classes and sampling examples from those classes that are partitioned into a {\em support} (i.e. training) and a {\em query} (i.e. test) set of examples. The model is tasked with training on the support set and is evaluated on its query set accuracy, finally the query set accuracies of many test episodes are averaged to obtain a measure of model performance on new learning problems. A detailed description of the setup can be found in~\cite{triantafillou2020metadataset}.

Meta-Dataset~\cite{triantafillou2020metadataset} is a large-scale few-shot classification benchmark that was introduced as a more realistic and challenging alternative to popular benchmarks such as mini-ImageNet \cite{vinyals2016matching}. While mini-ImageNet is constructed out of ImageNet classes (using 64, 16, and 20 classes to sample training, validation, and test episodes, respectively), Meta-Dataset is constructed out of many heterogeneous data sources whose classes are themselves partitioned into training, validation, and test sets of classes. Meta-Dataset, therefore, is a more challenging dataset in terms of robustness to distribution shift, which is compounded by the fact that two of its data sources (MSCOCO and Traffic Signs) are strictly reserved for test episodes (\autoref{fig:datasets} depicts Meta-Dataset examples).

SUR~\cite{dvornik2020selecting} tackles Meta-Dataset's domain heterogeneity by training separate backbones for each of the 8 data sources that define a training split of classes. Each backbone is trained to minimize classification error by sampling batches from it's corresponding dataset. However, validation for model selection of the backbone is performed by computing the classification error when it is used as a nearest centroid classifier (NCC). Finally during testing, all backbones are combined to form a single feature space. An inference procedure for test episodes uses the support set to select backbones to form a single representation, which is used to construct the final nearest-centroid classifier. 

Our experiments retrained SUR's 8 ResNet-18 backbones on their corresponding Meta-Dataset domain, and evaluated the model on test episodes following SUR's inference validation/test procedure, using the original open source codebase and hyper-parameters. Note that results in \autoref{table:sur_val} correspond to the validation set using the validation procedure, while \autoref{table:sur_test} report test results using the test procedure, outlined above. 

Finally we note that SUR's codebase is affected by a bug that causes the examples of each class to be visited in a deterministic order when sampling episodes. This bug was fixed in our experiments \footnote{\scriptsize\url{https://github.com/google-research/meta-dataset/issues/54}}. As a result both training and evaluation results were impacted, Traffic Sign test episodes are particularly affected as the deterministic order produces highly correlated episodes, which is why our reported baseline accuracies differ from those reported in the original SUR paper (with a larger margin noted on those from Traffic Signs).

\section{Experimental Results}

To investigate the effects of aliasing in the ResNet family of architectures we evaluate performance on ImageNet~\cite{ILSVRC15}, ImageNet-C~\cite{hendrycks2019robustness}, and Meta-Dataset~\cite{triantafillou2020metadataset}. Our aim for selecting these datasets is to highlight the difference in performance when evaluating in-distribution (ImageNet) versus out-of-distribution (ImageNet-C) or out-of-domain (Meta-Dataset) generalization. ImageNet training and validation datasets both consist of natural images with the same input resolution and compression format, therefore any features which may include aliasing artifacts learned during training are likely to transfer to the validation set. Finally, Meta-Dataset, a benchmark for few-shot image classification consists of eight training datasets and 10 test datasets. Eight of the the test datasets are considered in-domain because they correspond to one of the training datasets' domain (i.e. same source dataset), and two are considered out of domain (Traffic Signs and MSCOCO). Additionally, SUR evaluation is performed on three more test set domains, namely MNIST, CIFAR10, and CIFAR100.  


\begin{table*}[h]
\scriptsize
\begin{center}
\begin{tabular}{@{}lccccccccccccccc@{}}
  \toprule
  & \multicolumn{3}{c}{Blur before {\tiny (filter size $k$)}} & \phantom{a} & \multicolumn{3}{c}{Blur after {\tiny (filter size $k$)}} & \phantom{a} & \multicolumn{3}{c}{Blur both {\tiny (filter size $k$)}} & \phantom{a} & \multicolumn{3}{c}{Enlarge receptive field {\tiny (filter size $k$)}} \\ 
  \cmidrule{2-4} \cmidrule{6-8} \cmidrule{10-12} \cmidrule{14-16}
  Location & $k=3$ & $k=5$ & $k=7$ && $k=3$ & $k=5$ & $k=7$ && $k=3$ & $k=5$ & $k=7$ && $k=3$ & $k=5$ & $k=7$\\
  \midrule
  \DTLforeach{imagenet}{%
    \where=WHERE,%
    \bbthree=BB3,%
    \bbfive=BB5,%
    \bbseven=BB7,%
    \bathree=BA3,%
    \bafive=BA5,%
    \baseven=BA7,%
    \bboththree=BBOTH3,%
    \bbothfive=BBOTH5,%
    \bbothseven=BBOTH7,%
    \erfthree=ERF3,%
    \erffive=ERF5,%
    \erfseven=ERF7%
  }{%
    \DTLiffirstrow{}{\\}%
    \dtlformat{\where}&%
    \dtlformat{\bbthree}&%
    \dtlformat{\bbfive}&%
    \dtlformat{\bbseven}&&%
    \dtlformat{\bathree}&%
    \dtlformat{\bafive}&%
    \dtlformat{\baseven}&&%
    \dtlformat{\bboththree}&%
    \dtlformat{\bbothfive}&%
    \dtlformat{\bbothseven}&&%
    \dtlformat{\erfthree}&%
    \dtlformat{\erffive}&%
    \dtlformat{\erfseven}
  }
  \\ \bottomrule
\end{tabular}



\end{center}
\caption{\label{table:imagenet}
Imagenet results: rows demonstrate the impact of anti-aliasing the model's components from \autoref{fig:closer_look} individually (conv 1 is on the left, block-conv is in the middle, skip is on the right) 
while columns show blur variations from \autoref{fig:blur_variations}. Results show 
a significant accuracy increase when anti-aliasing the strided-skip connections of a Resnet-50.
Note that baseline accuracy is $76.49\%$. The values correspond to the mean accuracy over 3 runs with different seeds.
}
\end{table*}
{\setlength{\tabcolsep}{0.1em}
\begin{table}[ht]
\scriptsize
\begin{center}
\begin{tabular}{@{}ccccccccc@{}}
\toprule
\multicolumn{3}{c}{Blur filter location} & \phantom{a} & \multicolumn{2}{c}{Blur before {\tiny(filter size $k$)}} & \phantom{a} & \multicolumn{2}{c}{Blur after {\tiny(filter size $k$)}} \\
\cmidrule{1-3} \cmidrule{5-6} \cmidrule{8-9}
skip & max-pool & block-conv && $k=3$ & $k=5$ && $k=3$ & $k=5$ \\
\midrule
\checkmark &            &            && 78.55 {\tiny $\pm$ 0.05} & 78.65 {\tiny $\pm$ 0.13} && 78.54 {\tiny $\pm$ 0.03} & 78.60 {\tiny $\pm$ 0.17} \\
\checkmark & \checkmark &            && 78.68 {\tiny $\pm$ 0.14} & 78.65 {\tiny $\pm$ 0.07} && 78.61 {\tiny $\pm$ 0.14} & 78.57 {\tiny $\pm$ 0.10} \\
\checkmark &            & \checkmark && 78.92 {\tiny $\pm$ 0.02} & 78.84 {\tiny $\pm$ 0.04} && 78.87 {\tiny $\pm$ 0.07} & 78.94 {\tiny $\pm$ 0.02} \\
\checkmark & \checkmark & \checkmark && 78.85 {\tiny $\pm$ 0.13} & 78.80 {\tiny $\pm$ 0.14} && 78.85 {\tiny $\pm$ 0.02} & 78.72 {\tiny $\pm$ 0.03} \\
\midrule
\multicolumn{6}{l}{Baseline {\tiny (ResNet-50 + RandAugment + Swish, 180 epochs)}} & \multicolumn{3}{c}{77.38{\tiny $\pm$0.06}} \\
\bottomrule
\end{tabular}




\end{center}
\caption{\label{table:imagenet_iid_combinations_strong}ImageNet results when using data augmentation and smooth activations for various combinations of anti-aliasing locations. We see that all combinations reach higher accuracy than the baseline. Note that ``Blur before'' and ``Blur after'' only affect skip connections as anti-aliasing on max-pooling and block convolutions are fixed as ``Blur after''.
}
\end{table}
}

\subsection{ImageNet}

We used the official TensorFlow \cite{tensorflow2015-whitepaper} public code for training a ResNet-50 architecture, yielding a top-1 accuracy of $76.49\%$. The codebase reproduces the training pipeline and hyper-parameters from \cite{GoyalDGNWKTJH17}, in which models are trained for 90 epochs. 
\autoref{table:imagenet} shows the effect on top-1 accuracy of adding blur filters \emph{before}, \emph{after}, and \emph{before and after}, various operations in the network. Recall that adding blur kernels may also affect the back-propagation of gradients, as well as increase the receptive field. Our ablation studies are designed to disambiguate these effects. This is done by inserting blur kernels in each component of the ResNet model shown in Figure 1. \autoref{table:imagenet} reports accuracies as a result of adding a blur kernel before (77.14\%) or after (77.15\%) the skip connections \emph{that include sub-sampling}, and are immediately preceded by a trainable $1 \times 1$ convolutional layer. Note that inserting blur kernels for all convolutional layers (``all'' row) deteriorates the performance.  Anti-aliasing around the ``conv1'' layer degrades performance because it already has the capacity to learn a low-pass filter in its $7 \times 7$ kernel.  The last column, ERF (enlarged receptive field), is aimed to disambiguate whether the improved performance is truly due to anti-aliasing or merely the enlarged receptive field caused by adding the blur filter. This setting separates a striated convolution into sub-sampling followed by convolution without stride, and inserts the blur filter before sub-sampling. This gives the trainable convolutional layer access to a much larger receptive field, which actually causes the performance to degrade. Finally by combining anti-aliasing with data augmentation we are able to report a top-1 accuracy of 78.94\% \autoref{table:imagenet_iid_combinations_strong}, a competitive result on ImageNet-50, as compared to the known state-of-the art 79.01\% \cite{he2019bag} which proposes a significantly more complex approach. 

\subsection{ImageNet-C}


{\setlength{\tabcolsep}{0.1em}
\begin{table*}[ht]
\scriptsize
\begin{center}
\begin{tabular}{@{}lcccccccccccccccccccccccc@{}}
  \toprule
  & \multicolumn{3}{c}{Blur filter location} & \phantom{a} & \multicolumn{3}{c}{Noise} & \phantom{a} & \multicolumn{4}{c}{Blur} & \phantom{a} & \multicolumn{4}{c}{Weather} & \phantom{a} & \multicolumn{4}{c}{Digital} & mCE & Clean err.\\
  \cmidrule{2-4} \cmidrule{6-8} \cmidrule{10-13} \cmidrule{15-18} \cmidrule{20-23}
  Method & skip & max-pool & block-conv && Gauss. & Shot & Impulse && Defocus & Glass & Motion & Zoom && Snow & Frost & Fog & Bright && Contrast & Elastic & Pixel & JPEG && \\
  \midrule
  \DTLforeach{clean_corrupt_error}{%
    \method=Method,%
    \isskip=SKIP,%
    \ismaxpool=MAX-POOL,%
    \isblockconv=BLOCK-CONV,%
    \cleanerr=CLEAN,%
    \gausserr=GAUSS,%
    \shoterr=SHOT,%
    \impulseerr=IMPULSE,%
    \defocuserr=DEFOCUS,%
    \glasserr=GLASS,%
    \motionerr=MOTION,%
    \zoomerr=ZOOM,%
    \snowerr=SNOW,%
    \frosterr=FROST,%
    \fogerr=FOG,%
    \brighterr=BRIGHT,%
    \contrasterr=CONTRAST,%
    \elasticerr=ELASTIC,%
    \pixelerr=PIXEL,%
    \jpegerr=JPEG,%
    \mce=MCE%
  }{%
    \ifthenelse{\value{DTLrowi}=1}{%
    }{%
      \ifthenelse{\value{DTLrowi}=9}{\\\midrule}{\\}%
    }%
    \dtlformat{\method}&%
    \dtlformat{\isskip}&%
    \dtlformat{\ismaxpool}&%
    \dtlformat{\isblockconv}&&%
    \dtlformat{\gausserr}&%
    \dtlformat{\shoterr}&%
    \dtlformat{\impulseerr}&&%
    \dtlformat{\defocuserr}&%
    \dtlformat{\glasserr}&%
    \dtlformat{\motionerr}&%
    \dtlformat{\zoomerr}&&%
    \dtlformat{\snowerr}&%
    \dtlformat{\frosterr}&%
    \dtlformat{\fogerr}&%
    \dtlformat{\brighterr}&&%
    \dtlformat{\contrasterr}&%
    \dtlformat{\elasticerr}&%
    \dtlformat{\pixelerr}&%
    \dtlformat{\jpegerr}&%
    \dtlformat{\mce}&%
    \dtlformat{\cleanerr}%
  }
  \\ \bottomrule
\end{tabular}
\end{center}
\caption{\label{table:clean_corruption_error} Corruption Error (CE) on Imagenet-C corruptions, mCE, and Clean Error values when including our anti-aliasing variations. ResNet-50 and training for 90 epochs. Lower is better. We see that adding anti-aliasing decreases the errors on all corruptions except for Pixel and Blur. The errors were computed on the model achieving the median performance on ImageNet across 3 seeds. In our models, anti-aliasing is applied before the non linearities, as opposed to after as in \cite{zhang2019shiftinvar}.}
\end{table*}}

{\setlength{\tabcolsep}{0.1em}
\begin{table*}[ht]
\scriptsize
\begin{center}
\begin{tabular}{@{}lcccccccccccccccccccccccc@{}}
  \toprule
  & \multicolumn{3}{c}{Blur filter placement} & \phantom{a} & \multicolumn{3}{c}{Noise} & \phantom{a} & \multicolumn{4}{c}{Blur} & \phantom{a} & \multicolumn{4}{c}{Weather} & \phantom{a} & \multicolumn{4}{c}{Digital} & mCE & Clean err. \\
  \cmidrule{2-4} \cmidrule{6-8} \cmidrule{10-13} \cmidrule{15-18} \cmidrule{20-23}
  Method & skip & max-pool & block-conv && Gauss. & Shot & Impulse && Defocus & Glass & Motion & Zoom && Snow & Frost & Fog & Bright && Contrast & Elastic & Pixel & JPEG && \\
  \midrule
  \DTLforeach{data_aug_corruptions}{%
    \method=Method,%
    \isskip=SKIP,%
    \ismaxpool=MAX-POOL,%
    \isblockconv=BLOCK-CONV,%
    \cleanerr=CLEAN,%
    \gausserr=GAUSS,%
    \shoterr=SHOT,%
    \impulseerr=IMPULSE,%
    \defocuserr=DEFOCUS,%
    \glasserr=GLASS,%
    \motionerr=MOTION,%
    \zoomerr=ZOOM,%
    \snowerr=SNOW,%
    \frosterr=FROST,%
    \fogerr=FOG,%
    \brighterr=BRIGHT,%
    \contrasterr=CONTRAST,%
    \elasticerr=ELASTIC,%
    \pixelerr=PIXEL,%
    \jpegerr=JPEG,%
    \mce=MCE%
  }{%
    \ifthenelse{\value{DTLrowi}=1}{%
    }{%
      \ifthenelse{\value{DTLrowi}=6 \OR \value{DTLrowi}=12}{\\\midrule}{\\}%
    }%
    \dtlformat{\method}&%
    \dtlformat{\isskip}&%
    \dtlformat{\ismaxpool}&%
    \dtlformat{\isblockconv}&&%
    \dtlformat{\gausserr}&%
    \dtlformat{\shoterr}&%
    \dtlformat{\impulseerr}&&%
    \dtlformat{\defocuserr}&%
    \dtlformat{\glasserr}&%
    \dtlformat{\motionerr}&%
    \dtlformat{\zoomerr}&&%
    \dtlformat{\snowerr}&%
    \dtlformat{\frosterr}&%
    \dtlformat{\fogerr}&%
    \dtlformat{\brighterr}&&%
    \dtlformat{\contrasterr}&%
    \dtlformat{\elasticerr}&%
    \dtlformat{\pixelerr}&%
    \dtlformat{\jpegerr}&%
    \dtlformat{\mce}&%
    \dtlformat{\cleanerr}%
  }
  \\ \bottomrule
\end{tabular}
\end{center}
\caption{\label{table:data_aug_corruptions} Corruption Error (CE), mCE, and Clean Error values when including our anti-aliasing variations on top of ResNet-50 and training for 180 epochs with data augmentation. Adding anti-aliasing leads to a lower error than all existing models with the exception of ANT. ANT uses adversarial training and has an extra generative network, is significantly more expensive to train, has a higher clean error and has comparable Corruption Error to our simple modification. The errors were computed on the model achieving the median performance on ImageNet across 3 seeds. In our models, anti-aliasing is applied before the non linearities, as opposed to after as in \cite{zhang2019shiftinvar}.}
\end{table*}
}

\begin{figure}[ht]
\centering
\includegraphics[width=0.31\linewidth]{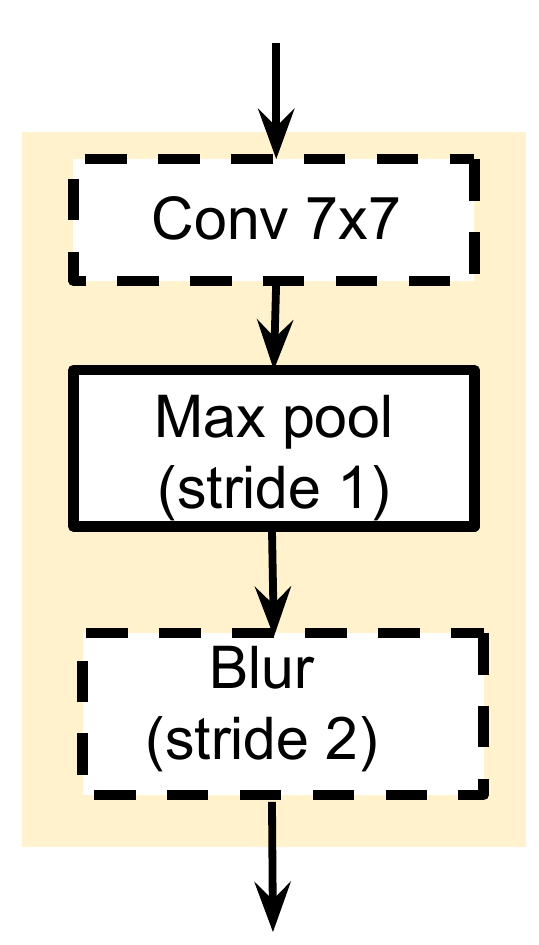}
\hfill
\includegraphics[width=0.68\linewidth]{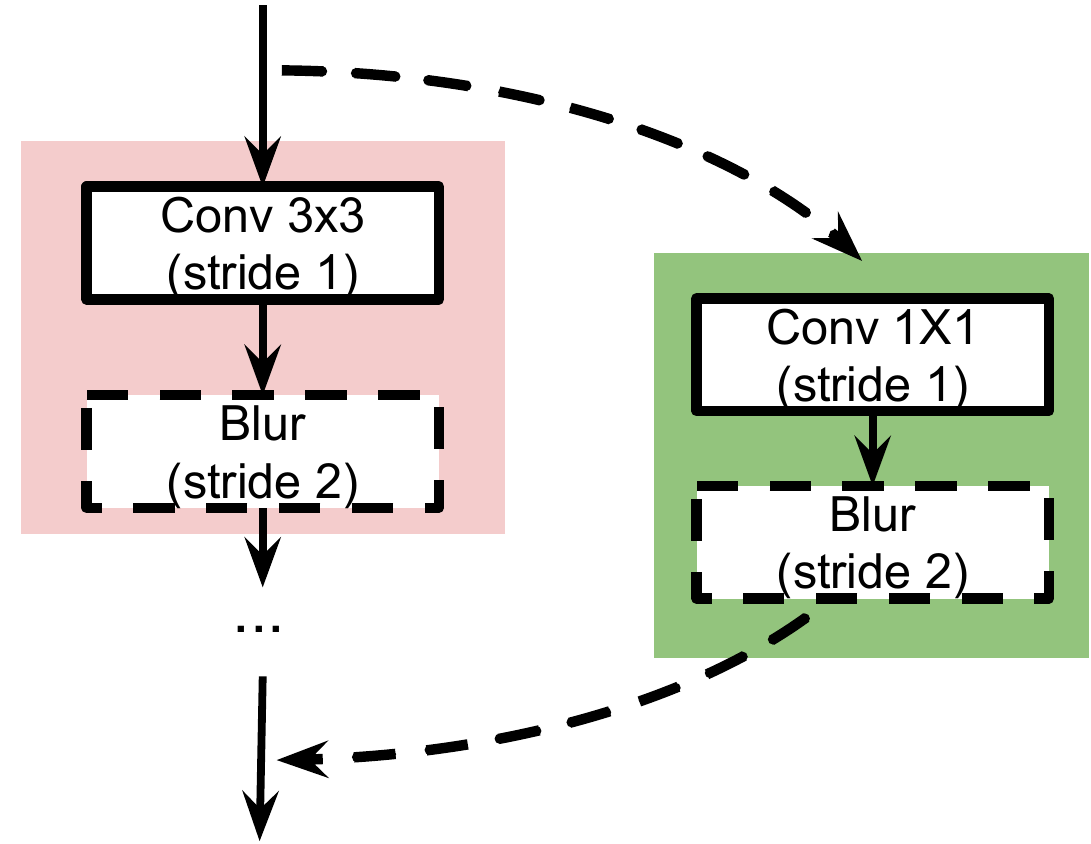}
\caption{The best variant on Imagenet-C (see \autoref{table:imagenet_iid_combinations_strong}). An analogous combination also produced the best results on Meta-Dataset. Adding ``Blur'' on strided-skip connections have the largest impact on performance.}
\label{fig:best_model}
\end{figure}

\autoref{table:clean_corruption_error} shows evaluation results for various architectures tested on ImageNet-C. The ``Clean err.'' colum shows the classifier's original top-1 error on the original ImageNet validation set. It also uses the corruption error measurement proposed by \cite{hendrycks2019robustness} that is defined as

\begin{equation}
\label{eq:crazy_error}
 CE^{f}_{c} = \frac{\sum^{5}_{s=1} E^{f}_{s,c}}{\sum^{5}_{s=1} E^{AlexNet}_{s,c}},
\end{equation}

where $E^{f}_{s,c}$ is the top-1 error of a classifier $f$ for a corruption $c$ with severity $s$. The mean Corruption Error (mCE) is taken by averaging over all the 15 corruptions. We adopted same values as \cite{hendrycks2019robustness} for  $E^{AlexNet}_{s,c}$. In summary, \autoref{table:clean_corruption_error} reports ordinary top-1 error in the ``Clean err.'' column and reports the error score defined in \autoref{eq:crazy_error} for all other columns. We used public code from \cite{cubuk2019randaugment} to replicate a data-augmentation baseline, but similar to \cite{hendrycks2020augmix}, we do not use augmentations such as contrast, color, brightness, sharpness, as they may overlap with the ImageNet-C test set corruptions. That is, the results build on top of the RandAugment code base, only uses 10 of their original 14 augmentations.
The published ResNet-50 baseline achieves 23.9\% on ImageNet-validation (``clean error'') and 76.7\% mCE on ImageNet-C. We are able to report significantly improved results of 23.5\% and 71.4\% using the same baseline architecture. We report the performance gains of various blur filter placements, suggesting that placing blur filters at all tested locations is the best variant, acheiving 23.0\% and 70.0\%. This represents a significant improvement over the results reported in \cite{zhang2019shiftinvar}. Next we show that performance can be further improved by adding smooth activation functions (Swish \cite{ramachandran2017searching}) and using additional data augmentation during training. \autoref{table:data_aug_corruptions} shows these results, all models were trained for 180 epochs in these experiments. We achieve a ``Clean'' top-1 error of 21.2\% and an mCE of 64.9\%. Note that unlike \cite{rusak2020simple} we demonstrate that mCE can be improved \emph{without sacrificing} clean error.

\subsection{Meta-Dataset with SUR}

\begin{table*}[ht]
\scriptsize
\begin{center}
\begin{tabular}{@{}lcccccccccccccc@{}}
  \toprule
  \multicolumn{2}{c}{} & SUR* & \multicolumn{3}{c}{Blur before} & \phantom{a} & \multicolumn{3}{c}{Blur after} & \phantom{a} & \multicolumn{3}{c}{ERF} & Stride 1 \\ 
  \cmidrule{4-6} \cmidrule{8-10} \cmidrule{12-14}
  Data source & Preprocessing && $k=3$ & $k=5$ & $k=7$ && $k=3$ & $k=5$ & $k=7$ && $k=3$ & $k=5$ & $k=7$ & \\
  \midrule
  \DTLforeach{meta_dataset}{%
    \source=DATASOURCE,%
    \preproc=PREPROCESSING,%
    \sur=SUR,%
    \bbthree=BB3,%
    \bbfive=BB5,%
    \bbseven=BB7,%
    \bathree=BA3,%
    \bafive=BA5,%
    \baseven=BA7,%
    \erfthree=ERF3,%
    \erffive=ERF5,%
    \erfseven=ERF7,%
    \strideone=STRIDE1%
  }{%
    \ifthenelse{\value{DTLrowi}=1}{%
        \dtlformat{\source} & \dtlformat{\preproc} &%
    }{%
      \ifthenelse{\value{DTLrowi}=9}{\\\midrule\multicolumn{2}{c}{\dtlformat{\source}} &}{\\\dtlformat{\source} & \dtlformat{\preproc} &}%
    }%
    \dtlformat{\sur}&%
    \dtlformat{\bbthree}&%
    \dtlformat{\bbfive}&%
    \dtlformat{\bbseven}&&%
    \dtlformat{\bathree}&%
    \dtlformat{\bafive}&%
    \dtlformat{\baseven}&&%
    \dtlformat{\erfthree}&%
    \dtlformat{\erffive}&%
    \dtlformat{\erfseven}&%
    \dtlformat{\strideone}
  }
  \\\bottomrule
\end{tabular}
\end{center}

\caption{\label{table:sur_val}Impact of the input resolution and of anti-aliasing on the initial convolutional layer before (Resblocks); Individual backbones work on input size $84\times84$. Scale pre-processing column indicates the relation between datasets original resolution and backbone input (up for up-scaled images and down for down-sampled images).
For each group, anti-alias was tested under the variations ``blur first'', ``blur second'', and ``enlarge receptive field'' (ERF), with filter sizes 3, 5 and 7. The results were compared to a standard ResNet-18 (SUR, first column), and to a ResNet-18 where the first convolution uses stride 1 (last column), i.e. no sub-sampling.
We see that anti-alias outperforms SUR for almost all datasets but cannot recover the performance of Stride 1, except on the two datasets which had little or no down-sampling (QuickDraw and Omniglot). 
}
\end{table*}

{\setlength{\tabcolsep}{0.4em}
\begin{table*}[ht]
\begin{center}
\scriptsize
\begin{tabular}{@{}lcccccccccccc@{}}
  \toprule
  & SUR* & \multicolumn{3}{c}{Blur skip + Conv1 w/stride 1} & GELU & \multicolumn{3}{c}{Blur all + Conv1 w/stride 1 + GELU } & \phantom{a} & \multicolumn{3}{c}{Blur skip + Conv1 w/stride 1 + GELU} \\ 
  \cmidrule{3-5} \cmidrule{7-9} \cmidrule{11-13}
  Data source && $k=3$ & $k=5$ & $k=7$ && $k=3$ & $k=5$ & $k=7$ && $k=3$ & $k=5$ & $k=7$\\
  \midrule
  \DTLforeach{meta_dataset_test}{%
    \source=DATASOURCE,%
    \sur=SUR,%
    \surs=SUR-S,%
    \aasthree=AAS3,%
    \aasthrees=AAS3-S,%
    \aasfive=AAS5,%
    \aasfives=AAS5-S,%
    \aasseven=AAS7,%
    \aassevens=AAS7-S,%
    \gelu=GELU,%
    \gelus=GELU-S,%
    \aasubgthree=AASUBG3,%
    \aasubgthrees=AASUBG3-S,%
    \aasubgfive=AASUBG5,%
    \aasubgfives=AASUBG5-S,%
    \aasubgseven=AASUBG7,%
    \aasubgsevens=AASUBG7-S,%
    \aasgthree=AASG3,%
    \aasgthrees=AASG3-S,%
    \aasgfive=AASG5,%
    \aasgfives=AASG5-S,%
    \aasgseven=AASG7,%
    \aasgsevens=AASG7-S%
  }{%
    \ifthenelse{\value{DTLrowi}=1}{%
    }{%
      \ifthenelse{\value{DTLrowi}=8 \OR \value{DTLrowi}=14}{\\\midrule}{\\}%
    }%
    \source&%
    \sur\ifthenelse{\value{DTLrowi}<14}{\tiny$\pm$\surs}{}&%
    \aasthree\ifthenelse{\value{DTLrowi}<14}{\tiny$\pm$\aasthrees}{}&%
    \aasfive\ifthenelse{\value{DTLrowi}<14}{\tiny$\pm$\aasfives}{}&%
    \aasseven\ifthenelse{\value{DTLrowi}<14}{\tiny$\pm$\aassevens}{}&%
    \gelu\ifthenelse{\value{DTLrowi}<14}{\tiny$\pm$\gelus}{}&%
    \aasubgthree\ifthenelse{\value{DTLrowi}<14}{\tiny$\pm$\aasubgthrees}{}&%
    \aasubgfive\ifthenelse{\value{DTLrowi}<14}{\tiny$\pm$\aasubgfives}{}&%
    \aasubgseven\ifthenelse{\value{DTLrowi}<14}{\tiny$\pm$\aasubgsevens}{}&&%
    \aasgthree\ifthenelse{\value{DTLrowi}<14}{\tiny$\pm$\aasgthrees}{}&%
    \aasgfive\ifthenelse{\value{DTLrowi}<14}{\tiny$\pm$\aasgfives}{}&%
    \aasgseven\ifthenelse{\value{DTLrowi}<14}{\tiny$\pm$\aasgsevens}{}%
  }
  \\\bottomrule
\end{tabular}
\end{center}
\caption{\label{table:sur_test} Evaluation of SUR models on 600 test episodes from MetaDataset. Columns: \emph{SUR} shows baseline performance using original  backbones.  \emph{Blur skip}: shows the effect of adding blur to skip connections with various blur kernel size ($k$), \emph{GELU}: replaces ReLU actions with GELU in all backbones, \emph{Blur all + GELU}: backbones includes blur filters at all subsampled modules, and GELU activations and finally \emph{Blur skip + GELU}: includes blur filters at strided-skip connections and GELU activations. 
Anti-aliased architectures used stride 1 in first convolutional layer.
Conclusions: adding blur at skip connections improves performance with or without GELU activations. GELU actions improve performance on their own. The best result is achieved by combining blur on skip connections with GELU activations.}
\end{table*}
}


SUR's preprocessing pipeline resizes the images of Meta-Dataset various datasets (``domains'') from their native resolutions to $84 \times 84$ using a bilinear interpolation \cite{dvornik2020selecting}. 
The experiments summarized in \autoref{table:sur_val} investigate the impact of this preprocessing on aliasing artifacts. 
The table reports \emph{validation} accuracies on each domain in Meta-Dataset. We chose to report validation accuracies for this experiment because SUR's validation procedure uses only the domain-specific backbones, i.e. without the confounding effects introduced by combining multiple backbones. The preprocessing column indicates the degree of downsampling (or upsampling) that was performed when constructing the data in each domain. We experiment with adding our anti-aliasing variants  to SUR's ResNet-18 first convolutional layer ($k$ indicates the size of the blur filter). The first convolutional layer of this architecture uses $5 \times 5$ kernels, and stride 2. \emph{This layer plays a particularly important role because we hypothesize that activations corresponding to images that were severely down-sampled are highly prone to aliasing via any further down-sampling.}
To test this, we also train an additional model with stride 1 in the first convolutional layer producing a controlled experiment that aims to isolate aliasing effects specifically in this layer (\autoref{table:sur_val} last column). Although adding a blur filter \emph{before} the first convolutional layer improves performance (from 75.27\% to 76.96\%), it is not as effective as removing stride on the first layer altogether (77.86\%). The conclusion we draw from this experiment is that although blur filters can mitigate aliasing effects of sub-sampling, they cannot preserve all information lost to this operation. Note that, avoiding sub-sampling altogether is a computationally impractical strategy (see \cite{DBLP:journals/jmlr/AzulayW19} for further discussion). Finally it can be seen that removing this first sub-sampling operation resulted in no performance improvement on datasets that were not severely downsampled during preprocessing, namely Omniglot and Quickdraw (\autoref{table:sur_val} last column).  

\autoref{table:sur_test} shows results of \emph{test} evaluation accuracies as a result of combining anti-aliasing with smooth GELU activation functions. Adding blur filters on the skip connections yields an average accuracy of 74.80\% (Blur skip + GELU). Including blur filters at all downsampling operations does not significantly improve average performance (74.82\%). We highlight that an average improvement of 3.75\% (absolute) was obtained (including a 2.73\% improvement on out-of-domain tasks). Note that this was achieved with only minor changes to the architecture while using the default hyper-parameters. 

\section{Conclusion}
Drawing from the classical sampling theorem from signal processing, we proposed simple architectural improvements to residual architectures to counter aliasing occurring at various stages of Residual Networks. These changes lead to substantial performance gains on both IID and OOD generalization, especially when combined with smooth activation functions and data augmentation.
Compared to other performance enhancement techniques, anti-aliasing is simple to implement, computationally inexpensive, and does not require additional trainable parameters. In all our experiments, we could not find a setting where it degraded the performance, making us recommend their use as a standard computational block in Residual Networks.






{\small
\bibliographystyle{ieee_fullname}
\bibliography{cvpr}
}

\end{document}